\documentclass[10pt,journal,compsoc]{IEEEtran}
\usepackage{amsmath,amsfonts}
\usepackage{algorithmic}
\usepackage{array}
\usepackage[caption=false,font=normalsize,labelfont=sf,textfont=sf]{subfig}
\usepackage{textcomp}
\usepackage{stfloats}
\usepackage{url}
\usepackage{verbatim}
\usepackage{graphicx}
\usepackage{graphicx}
\usepackage{amssymb}
\usepackage{booktabs}
\usepackage{color}
\usepackage{makecell}
\usepackage{float}
\usepackage{multirow}

\usepackage{balance}

\begin{document}

\title{Super-NeRF: View-consistent Detail Generation for NeRF super-resolution}
\author{Yuqi~Han, Tao~Yu, Xiaohang~Yu, Yuwang~Wang and Qionghai~Dai 
\IEEEcompsocitemizethanks{
\IEEEcompsocthanksitem The authors are
with Tsinghua University, Beijing 100084, China. \protect\\
E-mail: yqhan@mail.tsinghua.edu.cn,ytrock@126.com,\\yuxh21@mails.tsinghua.edu.cn, wang-yuwang@mail.tsinghua.edu.cn,\\
daiqionghai@tsinghua.edu.cn.
}
}

\markboth{IEEE TRANSACTIONS ON PATTERN ANALYSIS AND MACHINE INTELLIGENCE,~Vol.~xx, No.~xx, xx~2023}%
{Shell \MakeLowercase{\textit{et al.}}: Bare Demo of IEEEtran.cls for Computer Society Journals}

\IEEEtitleabstractindextext{
\begin{abstract}
The neural radiance field (NeRF) achieved remarkable success in modeling 3D scenes and synthesizing high-fidelity novel views. However, existing NeRF-based methods focus more on the make full use of the image resolution to generate novel views, but less considering the generation of details under the limited input resolution. In analogy to the extensive usage of image super-resolution, NeRF super-resolution is an effective way to generate the high-resolution implicit representation of 3D scenes and holds great potential applications. Up to now, such an important topic is still under-explored. In this paper, we propose a NeRF super-resolution method, named Super-NeRF, to generate high-resolution NeRF from only low-resolution inputs.  Given multi-view low-resolution images, Super-NeRF constructs a consistency-controlling super-resolution module to generate view-consistent high-resolution details for NeRF. Specifically, an optimizable latent code is introduced for each low-resolution input image to control the 2D super-resolution images to converge to the view-consistent output.  The latent codes of each low-resolution image are optimized synergistically with the target Super-NeRF representation to fully utilize the view consistency constraint inherent in NeRF construction. We verify the effectiveness of Super-NeRF on synthetic, real-world, and AI-generated NeRF datasets. Super-NeRF achieves state-of-the-art NeRF super-resolution performance on high-resolution detail generation and cross-view consistency. 
\end{abstract}

}
\maketitle
\IEEEdisplaynontitleabstractindextext
\IEEEpeerreviewmaketitle

\IEEEraisesectionheading{
\section{Introduction}}

Modeling 3D scenes for high-quality view synthesis has been a longstanding and crucial task in computer vision and graphics, which has various applications visual effects, and virtual and augmented reality~\cite{mescheder2019occupancy,kalantari2016learning,mildenhall2019local,li2021mine,flynn2019deepview}. 
Recently, NeRF~\cite{nerf} and its following works~\cite{codenerf,lin2021barf,barron2021mip,yu2021pixelnerf,martin2021nerf} have shown success in representing a 3D scene as a radiance field by modeling the light ray propagation from a 3D scene to a 2D image projection. 
As pointed out in~\cite{nerfsr}, NeRF implicitly models the physical inner view consistency and estimates the color and density in 3D space via the given 2D views. As a result, NeRF can render photo-realistic novel views given arbitrary capturing poses.  
However, as the quality of NeRF is highly limited by the resolution of the input view, for example, it is hard to create high-resolution NeRF with the collected low-resolution images.
Due to this physical modeling and the limitation of the MLP, the high-quality synthesis of NeRF highly relies on high-resolution views as input. 
Therefore,  the cost of capture, storage, and transmission of HR images could unavoidably increase.  

In analogy to the wide usage of image super-resolution (SR)~\cite{kim2010single,sun2008image,dong2014learning,esrgan,ExSR}, SR for NeRF is potentially an effective way to address the above issues. However, this important and practical topic is still under-explored. 
In this paper, we aim to achieve high-quality super-resolution for NeRF given low-resolution inputs only. 
A straightforward way is to conduct super-resolution on the low-resolution input views independently and then train a vanilla NeRF. However, the independent 2D image super-resolution ignores the 3D consistency across different views.
The deterministic inference of MLP makes NeRF sensitive to the inconsistent content of the input images. Hence, NeRF tends to take the average of multiple descriptions and further causes it to blur at the novel views. The cross-view consistent
super-resolution is the main challenge in NeRF super-resolution task.

\begin{figure*}[t]
  \centering
    \includegraphics[width=0.98\textwidth]{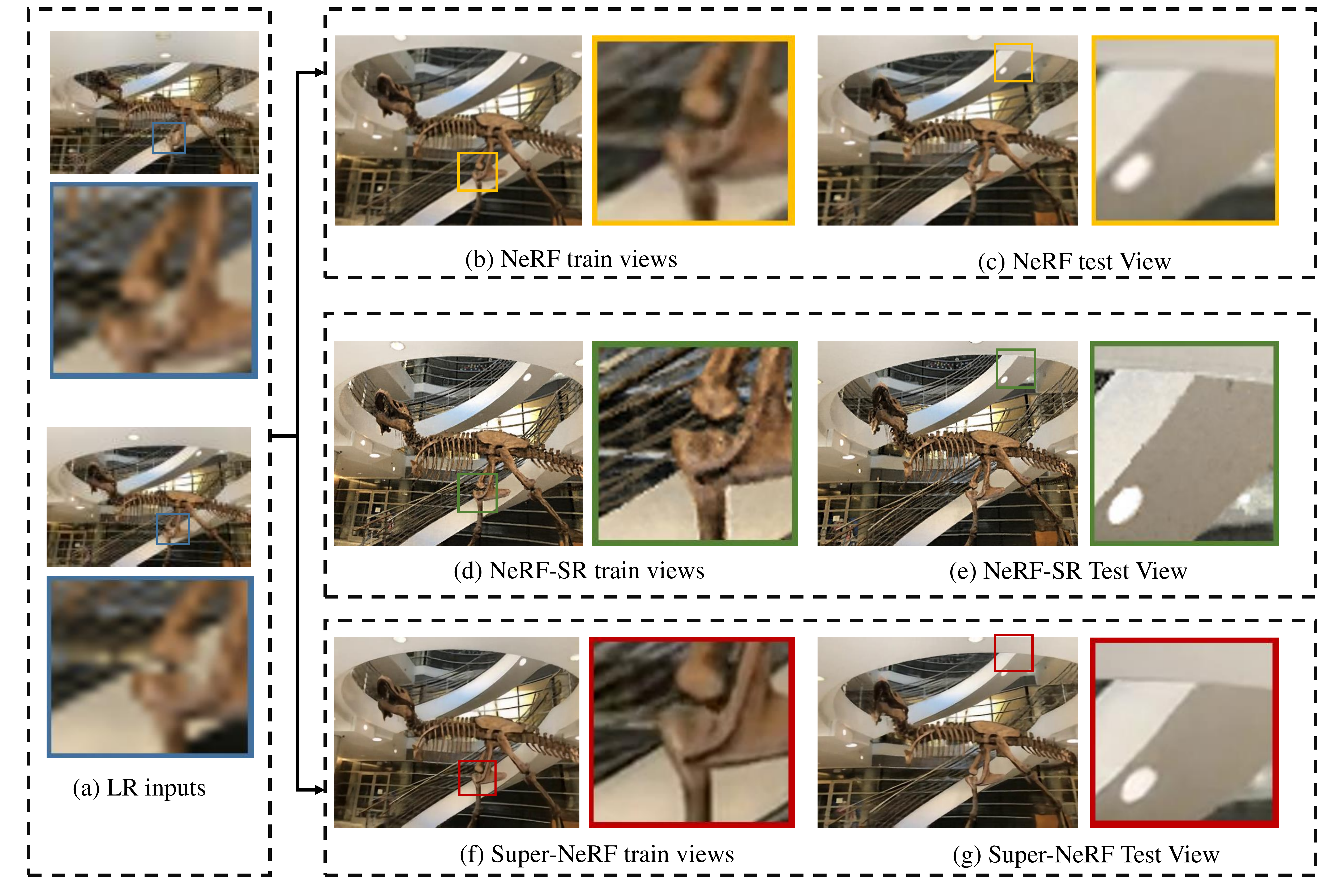}
  \caption{Comparison between Vanilla NeRF, NeRF-SR, and our Super-NeRF. With the input of extremely low resolutions (LR, around $100\times100$) inputs (a), the Vanilla NeRF directly uses them as train views (b) and generates test views with equal resolution (c). 
  NeRF-SR\cite{nerfsr} generates train views (d) and test views (e).
  The proposed Super-NeRF achieves view consistent 4x HR train views (f) and $4\times$ clean test view (g).}
  \label{fig:teaser}
\end{figure*}

To our best knowledge, the most related work is NeRF-SR~\cite{nerfsr}, which exploits the intrinsic 3D consistency embedded in NeRF to perform SR by a supersampling strategy on nearby views. 
However, NeRF-SR performs high-resolution reconstruction by exploiting the supersampling strategy to estimate color and density at the sub-pixel level rather than generating new high-resolution details to compensate for the low-resolution input.
Even with redundant low-resolution sampling, as SR is a non-linear ill-posed interpolation issue, the information provided by the low-resolution images may still be insufficient for high-quality SR. 
In this paper, to tackle this limitation, we propose a NeRF super-resolution method, named Super-NeRF, to construct high-resolution NeRF from low-resolution 2D image input. The result of vanilla NeRF, NeRF-SR, and Super-NeRF are shown in Fig.~\ref{fig:teaser}. NeRF-SR enforces the sampling to the sub-pixel, leading to the alias in the output. Super-NeRF can generate the 3D view consistent novel view synthesis with smooth and clear details.

We present a generative model-based SR without the high-resolution image guidance to derive the view consistent NeRF super-resolution. Specifically, we first introduce view-specific dense latent codes to control generating various SR results of images. On the basis of this, we propose a consistency-controlling super-resolution (CCSR) module to generate cross-view consistent 2D image SR results.
There are two objectives that CCSR needs to achieve for the view consistency NeRF super-resolution, i.e., the consistent detail generation in cross-view observation and high-fidelity HR detail generation. To address the first problem, we regard the NeRF as a constraint to maintaining cross-view consistency and proposing the mutual learning policy for CCSR and NeRF.  To address the second problem, we refer to a consistency enforcing module(CEM)\cite{ExSR} to make the 2D SR results conform to the LR images. 
We design two individual MLPs, denoting LR NeRF and HR NeRF, to accelerate the convergence. The LR NeRF is pretrained as coarse guidance of view consistency for image generation in the CCSR module. The HR NeRF is optimized with the CCSR concurrently to learn the view consistent detail information. We introduce a trade-off factor to conduct the weighted loss for LR NeRF and HR NeRF. As the iteration grows, the effect of LR NeRF on image generation gradually decreases, and the effect of HR NeRF on image generation gradually improves.

In the experiment, we validate the Super-NeRF in various radiance field construction scenarios, including planar light fields, 360-degree synthesized spherical light fields, 3D human faces, and even AI-generated light fields. Furthermore, we test the scalability of Super-NeRF in different camera settings, including input images with all LR, and input images with both LR and HR. The proposed Super-NeRF is not only robust to various scenes but achieves adaptive adjustment according to the camera settings. We believe that the proposed Super-NeRF has the potential to be extensively deployed in practical novel view synthesis tasks. 

Our main contributions can be summarized below:
\begin{itemize}
    \item{We propose Super-NeRF, a generative NeRF super-resolution method for view-consistent and high-resolution novel view synthesis with only LR images as input. As far as we know, we are the first to achieve a generative SR method for NeRF.}

    \item{We propose a consistency-controlling SR module, which locates the underdetermined attribute of the SR problem and successfully explores the SR solution space for generating details while maintaining multi-view consistency. }

    \item{Super-NeRF poses a strong generalization capacity for different kinds of light fields and naturally supports hybrid input resolution settings according to various practical requirements.}
\end{itemize}

In the following, we first briefly describe the related work in section~\ref{rw}. The framework of Super-NeRF and details of consistency controlling of SR is given in section~\ref{sr}. In section~\ref{ex}, we present a series of experimental results to verify the effectiveness of Super-NeRF. After that, we sum up the paper in section~\ref{co}.

\section{Related Works}
\label{rw}
In this section, we mainly review the works on the generative model-based  SR method and editable NeRF. We also briefly introduce the NeRF-SR\cite{nerfsr} at the end of the section. We thoroughly analyze the difference between NeRF-SR\cite{nerfsr} and Super-NeRF to make the contribution of the paper clear.
\subsection{Image SR}
Image SR is a classic topic in the computer vision research area. In this section, we only discuss the typical research achieving significant breakthroughs recently. 
Generative Adversarial Networks (GANs)\cite{goodfellow2020generative} have been shown to allow for photorealistic image SR based on the accurate perception of the natural images. The GAN-based SR solutions mainly take the LR images as the condition and adopt conditional-GAN~\cite{goodfellow2020generative,mirza2014conditional} framework to generate the SR output, such as~\cite{photo, esrgan, enhancenet, recovering}. The main objective of GAN-based SR is to generate realistic images that satisfy human's visual perception. 
The mentioned works dig into the statistical information of natural images and generate single results for the underdetermined SR problem.

The development of the diffusion model recently attracts extensive interest in 2D image generation. Specifically, the diffusion model-based SR solutions provide high-fidelity SR estimation by transforming the Gaussian noise in an HR representation with an LR condition. SrDiff\cite{SRDiff} adopts the UNet backbone as the approximation of the step-by-step iteration of the diffusion model and utilizes the RRDB architecture to achieve the LR encoding. However, SrDiff\cite{SRDiff} is difficult to train because of the long-term iteration and large dataset training. Diffir\cite{DiffIR} is proposed with a pretrained diffusion model and a fine-tuned diffusion model to improve the learning efficiency, which derives high-quality results during only a few interactions.  Sahak \emph{et al.} \cite{SRIW} focus on the remote blur and complex details of SR in the outside situation. Sahak  \emph{et al.} \cite{SRIW} formulates the real image quality degradation process and uses the results as the SR condition to achieve the self-supervised training, releasing the pressure from the large annotation dataset.  Recently, the proposition of Cascaded Diffusion Model\cite{CDiffM} performs high fidelity image generation, leading to 8x super-resolution from $32\times32$ to $256\times256$. By cascading the multiple diffusion models, Cascaded Diffusion Model\cite{CDiffM} enhances the diffusion ability with condition augmentation. The above research verifies the strong inference of the diffusion model. Currently, the diffusion model based methods little discuss the various SR details and controllable details generation for the SR task.

Some recent works~\cite{pulse,ExSR,SRFlow,deepsee,9837938} mention the ill-posed attribute of the SR problem and propose controllable SR solutions. Specifically, Menon  \emph{et al.}~\cite{pulse} performs the latent space exploration to avoid the image quality degradation caused by the pixel-wise average. 
They integrate the semantic segmentation probability maps to generate different textures in the different semantic classes.
Meanwhile, other works~\cite{ExSR,SRFlow} explore the solution space of the SR task and build different SR results by adding different conditions to LR images. It is worth mentioning that Bahat et al.~\cite{ExSR} analyzes the correspondence between the SR controlling latent code and LR images, leading to the explorable SR for the 2D LR images.

The mentioned works discuss the solution of the SR task but do not extend the task to 3D space. So far, there is no research discussing the extension of 2D SR to 3D SR with GAN-based solutions. In this paper, we borrow the thought of 2D SR exploration, providing possible solutions for constructing 3D consistent SR for NeRF.

\subsection{Editable NeRF}

After the proposition of NeRF\cite{nerf}, in the past two years, there are a lot of works focusing on the editing of NeRF.  By disentangling objects in the 3D space separately, NeRF can be formulated into various shapes and appearances. 
For example, Jang  \emph{et al.}.~\cite{codenerf} renders the unseen textures via control of the latent code. Other works~\cite{graf,giraffe,nerfediting,chan2021pi} accomplish the simple domination of appearance to the 3D objects, achieving the simple transformation of a single object. Hao  \emph{et al.}~\cite{gancraft} integrates a semantic block to generate corresponding 3D objects. GSN~\cite{GSN} performs the decomposition of 3D scenes and enables editing for each object. Some work \cite{nerfies,tretschk2021non,pumarola2021d,park2021hypernerf} introduce deforming scene radiance field construction to perform accurate implicit representation of natural scenes.

Different from the above-mentioned object editing task, some research proposes the editing of the whole 3D scene. Gu  \emph{et al.}~\cite{stylenerf} constructs a generator to achieve detailed consistency for the global scene. Some research ~\cite{stylizednerf,nguyen2022snerf} conducts the style transfer from the 2D image to the NeRF, maintaining consistency at the same time. Chen  \emph{et al.}~\cite{chen2022upst} accomplishes the universal photorealistic style transfer with accurate view consistency.

Up to now, the editing of NeRF depends on object structure or scene distribution. However, it is difficult to capture the shape or distribution of the details in NeRF super-resolution task. The above research cannot be directly used as the solution for NeRF super-resolution.

\subsection{SR of NeRF}

Currently, the 3D view consistent SR for the implicit representation is little researched. The NeRF backbone still needs the HR inputs to synthesize the HR novel views. We believe that the 3D view consistent SR of NeRF is a potential solution to support the implementation of NeRF in practical applications.

NeRF-SR\cite{nerfsr} recovers the high-frequency details in a sub-pixel manner, which forces the values of LR pixel to equal the mean value of HR sub-pixels. By reversely solving the linear formulation from LR to HR, NeRF-SR generates the novel view super-resolution approximate to the HR ground truth. However, there are two weaknesses to NeRF-SR. First, the linear formulation from LR to HR is too simple to generate aliases in the edges. Second, the high-quality super-resolution of eRF-SR relies on the HR refinement solution by per-patch warping to guide the network inference.

Different from NeRF-SR, the proposed Super-NeRF mainly focuses on solving the super-resolution of NeRF in a more challenging scenario, i.e., generating plausible 3D consistent HR details based on less information(LR observations only). Super-NeRF explores the SR solution space to acquire a reasonable solution, which satisfies the perception requirement of the human eyes and the consistency of 3D views. Moreover, Super-NeRF does not rely on any HR ground truth for high-quality HR details.

\section{Super-NeRF}
\label{sr}
In this section, we first introduce and analyze the framework of Super-NeRF, followed by detailed descriptions of the proposed CCSR module and the mutual learning strategy for NeRF super-resolution. Finally, we introduce the design of the loss function.

\begin{figure*}[t]
  \centering
  \includegraphics[width=0.95\textwidth]
  {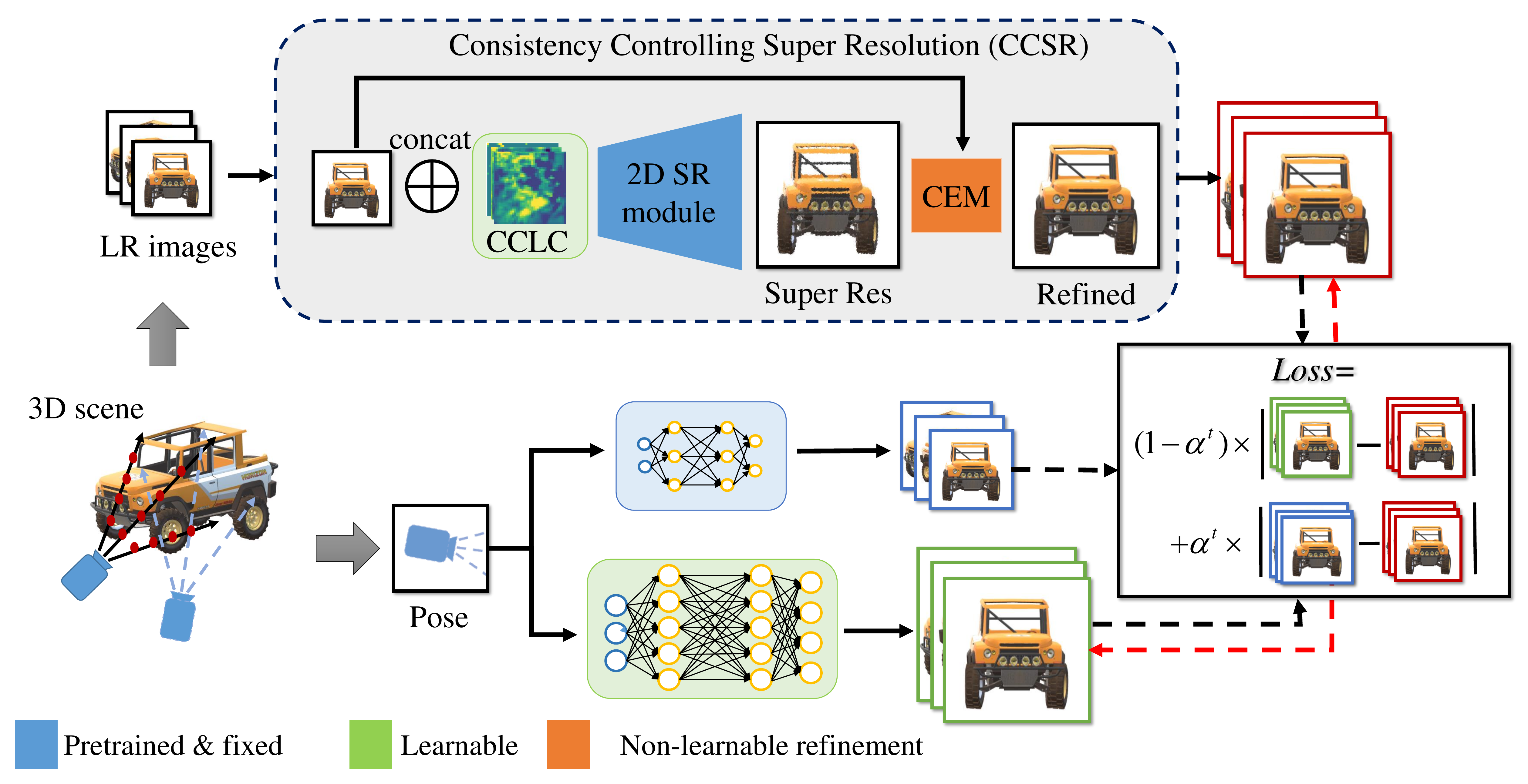}
  \caption{Super-NeRF framework: with a pretrained LR NeRF and a consistency-controlling SR module, Super-NeRF optimizes the HR NeRF and corresponding SR latent code simultaneously, and finally achieves view-consistent HR novel view synthesis. The CCLC represents the consistency-controlling latent code.}
  \label{fig:stru}
\end{figure*}


\subsection{Framework}
The framework of Super-NeRF is shown in Fig.~\ref{fig:stru}. 
Super-NeRF consists of a CCSR module, an LR NeRF, and an HR NeRF. 
The CCSR module is responsible for exploring various image SR solutions for each LR input. The HR NeRF constructs the implicit representation of the volume rendering based on the output of CCSR. The CCSR and the HR NeRF perform mutual learning and converge to the view-consistent NeRF super-resolution. In other words, after joint training, the CCSR  generates view-consistent 2D SR results, and HR NeRF is constructed with the output of CCSR. 

The LR NeRF is constructed by the LR inputs in advance as a warm-up for keeping view consistency and accelerating HR detail generation as well. 
The LR NeRF is lightweight (only a quarter the size of the HR NeRF) and can be efficiently pre-trained using the input LR images. The two advantages of the pre-trained LR NeRF in our pipeline are: i) it provides a strong view-consistency constraint to the output of the CCSR module (Even though the HR NeRF is not mature in the early stages of Super-NeRF training, the CCSR can still maintain coarse view-consistency.), and ii) it provides a stable and effective basis for the HR NeRF, and the reconstruction efficiency can be extremely improved. 

The trade-off between sufficient sampling and efficient learning is considered here. We set the LR NeRF and HR NeRF with different positional encoding dimensions. Specifically, we use lower-dimension positional encoding in LR NeRF and higher-dimension positional encoding in HR NeRF to capture the different scales of the details in the input images with different resolutions. 



\subsection{Consistency Controlling SR Module (CCSR)}

The main goal of the CCSR module is to generate controllable and plausible HR details given the LR input. Based on this, Super-NeRF can find the group of high-quality SR results satisfying the view-consistent constraint as the training data of NeRF.
Basically, the CCSR module is composed of a consistent controlling latent code (CCLC), a pre-trained SR module, and a CEM module. The CCLC is designed to control the various HR detail generation. The pre-trained SR module generates highly perceptible HR results to the human eye. The CEM maintains the consistency between the generated HR result and LR input.

To fulfill view-consistent HR detail generation across multiple viewpoints for general scenarios, the 2D SR backbone of the CCSR module should not only have strong SR generalization ability for handling different scenarios but more importantly, has the potential to provide enough DoF for controlling the generated details. Thus, we design the CCLC to control various detail generation. 
The CCLC is required to explore the SR results as much as possible, so the chance of finding the cross-view consistent SR results is increased.
Hence, we leverage the per-pixel control for 2D image SR to explore the solution space in super-resolution comprehensively.
The CCLC is designed based on the LR image size as well as the multiplier of the image enlargement.
For each LR image with $H\times W$ pixels, the CCLC is composed of a $3\times SH\times SW$-dimension latent code, in which $3$ dimension corresponds to the image channel. In this paper, we use the $4\times$ super-resolution backbone, so we set $s=4$ in the following description.

We adopt the ESRGAN, a generative SR method, as our 2D SR backbone, which is capable of generating HR details for general scenes. 
However, the generated details of standard ESRGAN depend on the feature of the actual training dataset, and the standard ESRGAN cannot generate various HR details for the same LR input.
In this case, even though we observe the same objective from a different observing viewpoint, the standard ESRGAN might generate inconsistent details.
If we directly input inconsistent SR results to the NeRF, the NeRF construction will be failed.
We concatenate the LR input to the random CCLC during training to guide the ESRGAN to generate various results of super-resolution. In the training process, the random CCLC  focuses on traversing the HR details space of plausible images. After training, an arbitrary latent code corresponds to an identified HR details generation. 

Except for the standard ESRGAN, the CCSR module includes a Consistency Enforcing Module (CEM)\cite{ExSR} to remove the artifact caused by SR Network as much as possible.
Given the SR output $\hat C_{HR, i}$ and the blur kernel $H$, the objective of CEM is to find a solution that perfectly matches the LR image $ C_{LR, i}$ after blurring and close to the super-resolution output as well, i.e.,
\begin{equation}
\begin{split}
    &\arg\min |\hat C_{HR, i} - C_{HR, i}|,\\
    &\quad s.t.\quad H C_{HR, i} =  C_{LR, i}.
    \end{split}
\end{equation}

The optimization details are shown as \cite{ExSR}. The output of CEM is can be seen as the composition of the LR input and the ESRGAN output. For the output of CCSR $C_{HR, i}$, considering the constraint of Eq.(1), we first map LR content to SR result with the orthogonal projection matrix, denoted as $H^T(HH^{-1}H)$, $H^T(HH^{-1}H)C_{HR, i} =H^T(HH^{-1}) C_{LR, i}$. Later, the rest information is from the ESRGAN output, which is mapped by the matrix $I -H^T(HH^{-1}H)$.
Hence, the output of the SR module is adjusted as
\begin{equation}
    \begin{split}
        C_{HR, i}  &= (I -H^T(HH^{-1}H)\hat C_{HR, i}\\
        &+ H^T(HH^{-1}) C_{LR, i}.
    \end{split}
\end{equation}

With the CEM, the output of CCSR is always conforming to the LR input. The CCLC is guided to control the generation of details instead of the whole picture, thus ensuring more controllable details and multi-view consistency.

It is noted that the CEM optimization can be implemented after any SR backbones. The solution of CEM is finally related to the LR input, SR output, and the blurring process.

\subsection{Loss Functions}
We adopt the mutual learning strategy in this paper to construct a joint optimization between the CCSR module and the HR NeRF. Both the LR NeRF and HR NeRF provide view-consistency constraints and the CCSR module generates various HR details. Based on the mutual learning scheme, NeRFs help the CCSR module to find the HR details satisfying the cross-view consistency. Meanwhile, the outputs of CCSR for different views are the inputs to construct HR NeRF. The loss is composed of the mean square error (MSE) between the output of the CCSR module and the NeRFs. 

At the beginning of the optimization, the HR NeRF has little knowledge to construct the correct consistency for CCSR. Hence, we use the pre-trained LR NeRF as a relaxed consistency constraint. We introduce a tradeoff function $\alpha$ here, ranging from $0$ to $1$, which is gradually decreasing with the accumulation of iteration times.
Based on the above description, the loss of mutual learning of viewpoint $i$ is defined by 
\begin{equation}
\begin{split}
\mathcal{L}_{SR} &= \alpha(t) * |C_{LN,i} - C_{HR,i}|\\
&+ (1-\alpha(t) )\times |C_{HN,i} - C_{HR,i}|
\end{split}
\end{equation}
where $t$ denotes the iteration times during the mutual learning, $\alpha(0) = 1, \lim_{t\to\infty} \alpha(t) = 0$.

We also define a penalty loss to avoid the output color of CCSR exceeding the valid range, denoted as $\mathcal{L}_{range}$. The 
\begin{equation}
\mathcal{L}_{range} = \frac{1}{N}\sum_{x}|\hat x- \text{truncate}_{0,1}(\hat x)|_1.
\end{equation}
where $N$ indicates the number of pixels, and 
\begin{equation}
    \text{truncate}_{0,1}(\hat x)=\left\{
\begin{aligned}
 0 &\text{ if }\hat x<0 ,\\
 1&\text{ if }\hat x>1 ,\\
\hat x&\text{ otherwise. }
\end{aligned}
\right.
\end{equation}

In summary, the final loss $\mathcal{L}$ for Super-NeRF training is defined as
\begin{equation}
    \mathcal{L} = \mathcal{L}_{SR} + \mathcal{L}_{range}.
\end{equation}

\begin{figure*}[t]
  \centering
  \includegraphics[width=0.9\textwidth]{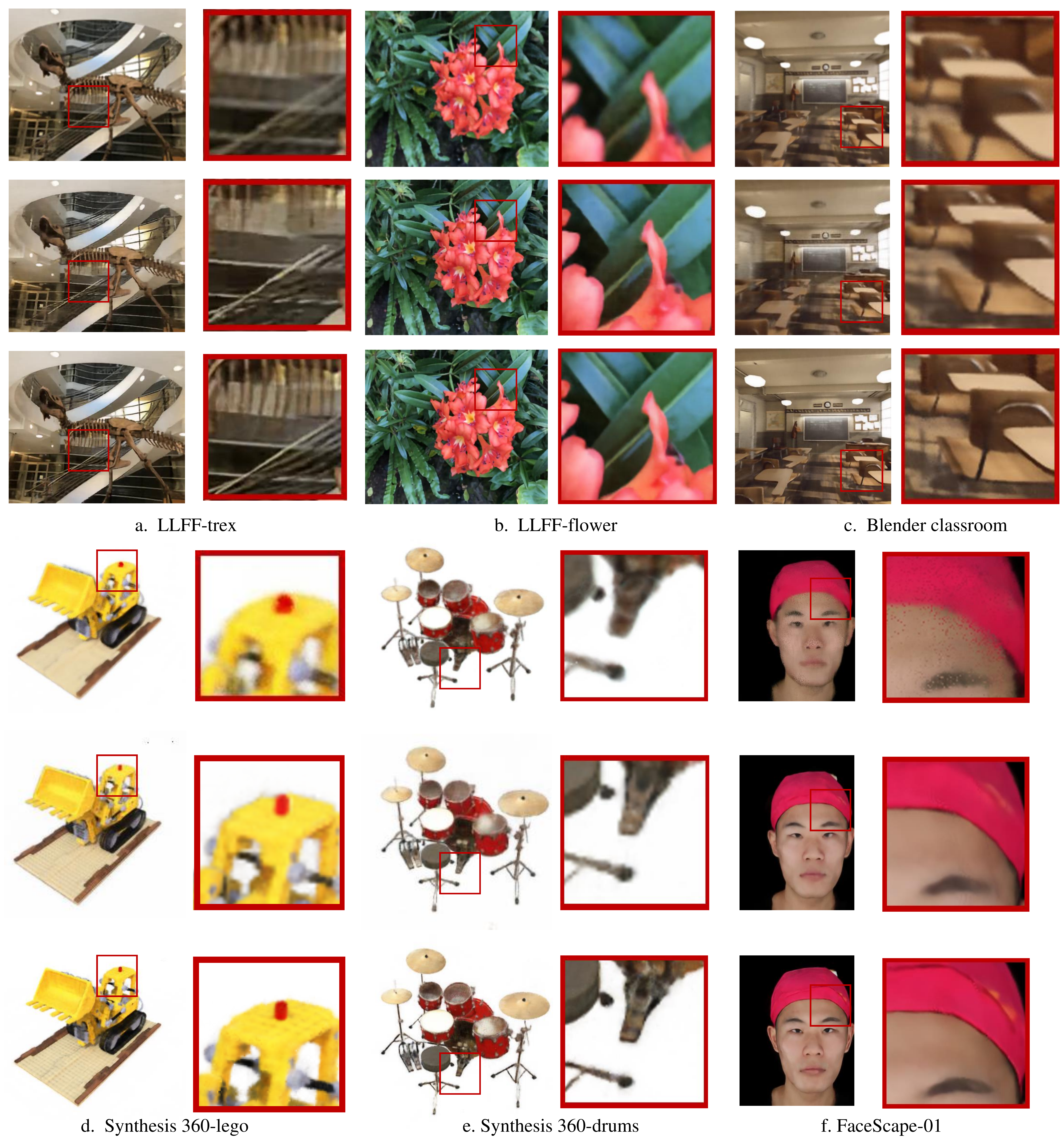}
  \caption{Comparison with the LR NeRF, the baseline method ESR-NeRF (using independently SR-ed images for HR NeRF training directly), and our Super-NeRF on various datasets ranging from complex scenes to human faces. For each case, from top to bottom are LR-NeRF, ESR-NeRF, and Super-NeRF respectively. Given low-resolution images only, Super-NeRF achieves $4\times$ SR of the neural radiance fields with well-generated details. Please see the supplementary video for high-resolution view-consistent novel view synthesis results generated by Super-NeRF.}
  \label{fig:result1}
\end{figure*}

\noindent\textbf{Implemetation details} {
We first pre-train the model of the LR NeRF and the 2D image super-resolution backbone and fix the two modules during the training of Super-NeRF.
For each iteration of the training process, we randomly choose a training viewpoint $i$ and super-resolve LR image $C_{LR,i}$ with CCSR to SR result, denoting as $\hat C_{HR,i}$. The $\hat C_{HR,i}$ is refined by the CEM module  and  the final output of CCSR is denoted as $C_{HR,i}$. According to the pose of the $i$, we upsample the rays and send the points along the ray to LR NeRF to acquire an image the same size as $C_{HR,i}$, denoting as $C_{LN,i}$. We send the same queried rays to the HR NeRF and generate the image output $C_{HN, i}$. The loss is constructed by $C_{HR,i}$, $C_{LN,i}$, and  $C_{HN, i}$. 
The HR NeRF and CCLC are trained via the mutual learning strategy and finally converge to a cross-view consistent NeRF super-resolution result.
Once the Super-NeRF finishes training, the HR NeRF can implement the novel view synthesis for any viewpoint. 

For all the experiments, we use a pretrained ESRGAN \cite{ExSR} as the backbone of 2D image SR in the CCSR module. We choose a $4\times$ SR pre-trained model provided by \cite{ExSR}. The code is implemented on a single NVIDIA RTX 3090 GPU. 
The average training time of the LR NeRF and the Super-NeRF are 30 minutes and 10 hours, respectively. 
}

\section{Experiments}
\label{ex}
In this section, we report a series of qualitative and quantitative evaluations to verify the effectiveness of the proposed method. Moreover, we discuss the ablation study  with different experiment settings. Finally, we validate the generality of the proposed Super-NeRF to extend the applications.

\begin{figure*}[t]
  \centering
  \includegraphics[width=0.95\textwidth]{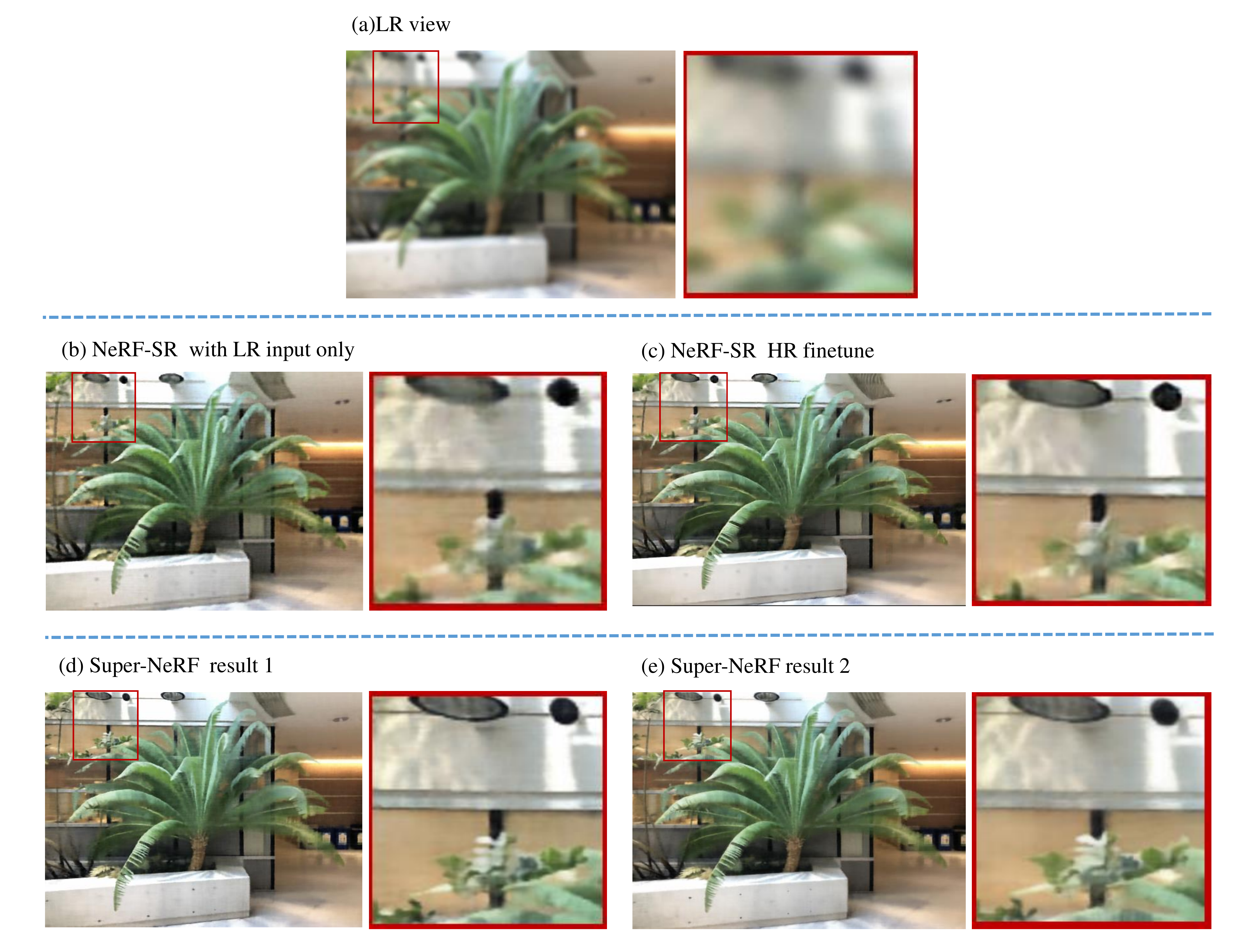}
  \caption{Comparison with NeRF-SR. From left to right: LR inputs, NeRF-SR without HR ground truth fine-tune, NeRF-SR with HR ground truth fine-tune, two results of Super-NeRF(with different generated details).}
  \label{fig:result2}
\end{figure*}

\subsection{Experimental settings}
In the qualitative evaluation, we compare Super-NeRF with 4 baselines. We first  use the pretrained LR NeRF directly for querying rays for HR pixels. For the second one, the LR images are first fed to the ESRGAN individually, and the SR results are input to NeRF, which we named ESR-NeRF. We conduct an individual comparison with NeRF-SR\cite{nerfsr}, which achieves the SR of NeRF by super-sampling the scenario.
We show the performance of NeRF-SR with HR ground truth finetune to further prove the performance of Super-NeRF. Moreover, to verify the exploration capability of Super-NeRF, we show different HR details generated by Super-NeRF in the same scenario.

We use 4 different datasets to conduct the qualitative experiments, including the LLFF dataset~\cite{llff} with real-captured planer light fields, the Synthesis 360 dataset with synthesized 360 light fields~\cite{nerf}, the Blender dataset~\cite{neff2021donerf} with complex scenes like classrooms, and the FaceScape dataset~\cite{facescape} which contains real-captured multi-view human face images. The
performance across different datasets can demonstrate
the strong generalization capacity of Super-NeRF.

\subsection{Metrics}
Since Super-NeRF only takes LR images as input, we can not guarantee the generated
HR details are consistent with the ground truth HR images. Hence, we do not compare the PSNR or SSIm with HR ground truth.  In the Quantitative evaluation, we investigate the quality of the generated details. We follow \cite{LearntoStylenv, stylenerf} to regard LPIPS as the view consistency investigator of the generated radiance fields. 
Given two images, the LPIPS can reflect the similarity between them in human perception level, in which lower values denote the higher similarity.
After that, we investigate the perception evaluation metrics NIQE, which reflects the acceptable grade of the generated image details to the human visual perception. With a lower value of NIQE, we regard the image  as more acceptable to human visual perception, i.e., higher quality.



\subsection{Qualitative Comparision on NeRF super-resolution}
The qualitative comparison of four datasets is shown in Fig.~\ref{fig:result1}. We can see the proposed Super-NeRF demonstrates more clear edges and texture details. It is worth noting that according to Fig.~\ref{fig:result1} (the lego case), Super-NeRF generates a texture different from the ground-truth image. This phenomenon verifies that Super-NeRF indeed provides a solution to the under-determined SR problem by creating HR detail, rather than reconstructing details exactly consistent with the HR ground truth. 

Fig.~\ref{fig:result2} presents the comparison with NeRF-SR. Fig.~\ref{fig:result2}(a) shows the LR views of the LLFF fern scene. Fig.~\ref{fig:result2}(b)(c) presents the results of NeRF super-resolution, which displays a large gain in image quality with HR ground truth . The different results of Super-NeRF are presented in Fig.~\ref{fig:result2}(d)(e). According to Fig.~\ref{fig:result2}, we can see that Super-NeRF not only performs better than NeRF-SR but generates various plausible SR results.

\begin{figure*}[t]
\centering
  \includegraphics[width=\textwidth]{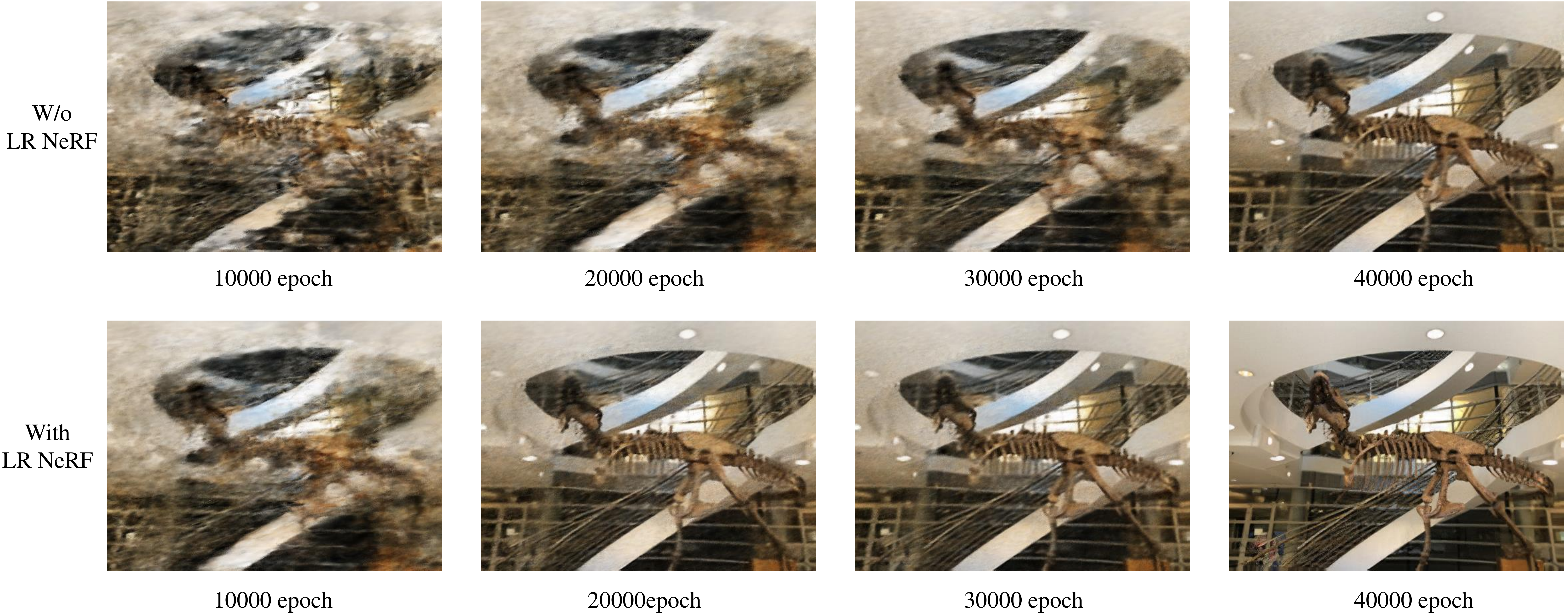}
  \caption{Ablation experiment on LR NeRF supervision considering the convergence speed of Super-NeRF, up: without LR NeRF, down: with LR NeRF. }
  \label{fig:abl2}
\end{figure*}

\subsection{Quantitative Comparison on NeRF super-resolution}

Following the NeRF stylization works~\cite{stylizednerf}, we use the LPIPS to evaluate the performance of the proposed solution. 
Specifically, we warp a novel view $i$ to another view $j$ with estimated depth expectation and calculate the LPIPS~\cite{LPIPS} between the raw view and warped view. 
The LPIPS is investigated in different scenes and different disparities for the same point in different views. 
Moreover, we refer to \cite{NIQE}, an image perception quality evaluation metric, to indicate the perceiving quality of the Super-NeRF.  
The results are shown in Tab.~\ref{tab:quan_view_consistency}. Compared to the ESR-NeRF, LR NeRF, and NeRF-SR, we see that the Super-NeRF reduces LPIPS and NIQE by a large margin, indicating better perception level and view consistency. The proposed CCSR generates a reasonable output for each 2D image and NeRF strongly constrains the view-consistent views generation. 
Also, we introduce the results of the HR NeRF and NeRF-SR with ground truth finetuned. According to Tab.~\ref{tab:quan_view_consistency}, the proposed Super-NeRF achieves comparable results with the HR ground truth, further indicating the effectiveness of Super-NeRF in the NeRF super-resolution task.

\begin{table}[htbp]
\caption{ Comparison on NIQE and LPIPS.} 
\renewcommand{\arraystretch}{1.2}
\begin{tabular}{c|cc|ccc}
\toprule
\multirow{2.5}{*}{Methods} &\multicolumn{2}{c}{NIQE$\downarrow$}&\multicolumn{3}{c}{LPIPS$\downarrow$}  \\
\cmidrule{2-6}
 & LLFF & Syn360& 3 pix & 10 pix& 15 pix \\
\midrule
LR-NeRF & 10.129 & 10.545&{0.0571}&0.0880&{0.1072}  \\
ESR-NeRF &6.018&6.227&0.0701&0.1202&0.1627  \\
NeRF-SR &5.736&6.210&0.0746&0.1403&0.1884  \\
Super-NeRF &{\color{red}\textbf{4.457}}&{\color{red}\textbf{5.136}}&{\color{red}\textbf{0.0550}}&{\color{red}\textbf{0.0873}}&{\color{red}\textbf{0.1064}}  \\
\midrule
\makecell{NeRF-SR\\(HR finetuned)} &{\textbf{4.640}}&{\textbf{5.043}}&{\textbf{0.0626}}&{\textbf{0.1011}}&{\textbf{0.1107}} \\
HR NeRF &{\color{blue}\textbf{4.218}}&{\color{blue}\textbf{5.027}}&{\color{blue}\textbf{0.0479}}&{\color{blue}\textbf{0.0821}}&{\color{blue}\textbf{0.0927}}  \\
\bottomrule
\end{tabular}
\label{tab:quan_view_consistency}
\end{table}


\begin{figure}[t]
\centering
  \includegraphics[width=0.45\textwidth]{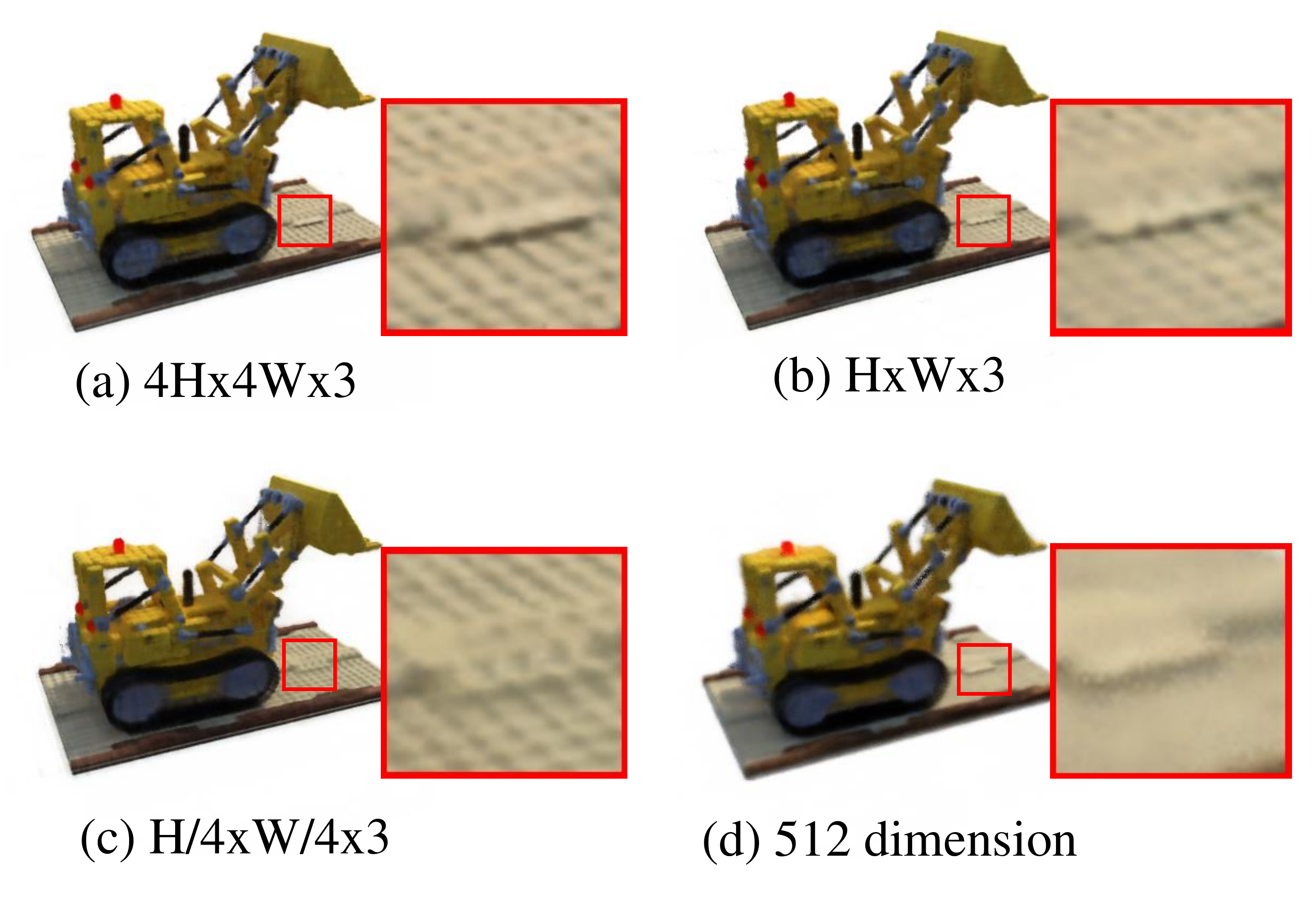}
  \caption{Qualitative evaluation on latent code. A larger size of latent code leads to better SR, and (d)  cannot achieve detailed control and generation.}
  \label{fig:abl}
\end{figure}
\subsection{Ablation Study}
\noindent\textbf{Evaluation of LR NeRF.}
As the LR NeRF is a module to provide basic constraints, we evaluate the effectiveness of the pre-trained LR NeRF in the whole framework. We conduct a comparison between the framework with and without LR NeRF, respectively. According to Fig.~\ref{fig:abl2}, the CCSR with LR NeRF supervision converges much faster than the one without LR NeRF. This demonstrates the effectiveness of our training strategy and the LR NeRF supervision for the radiance field SR tasks. 

\noindent\textbf{Evaluation of latent code dimension.}
As the latent code plays an important role in understanding how the SR works, we first conduct an ablation study discussing the impact of the latent code. 
We set up three baselines. The first one is a 512 dimension setting, following StyleNerf\cite{stylenerf}, which is mapped into Surper-NeRF by an MLP.
And other two are downsampling the raw size of CCLC with $1/4\times$ and $1/16\times$ respectively.

According to the results shown in Fig.~\ref{fig:abl}, different dimensions of the latent code lead to the different granularity of details. 
With the decrease in the latent code size, we can see that the bulge on the floor is gradually inconspicuous. This demonstrates that the latent code can be downsampled in scenes that are not much sensitive to details. 
However, using a latent code of 512 dimensions for NeRF super-resolution in the whole scene will severely deteriorate the overall performance. This indicates that the NeRF super-resolution task is different from the stylization task, such as StyleNerf\cite{stylenerf}, and SR task requires higher DoF control of the generated details. 

\noindent\textbf{Evaluation of Hybrid Resolution.}
A new ablation experiment was designed to validate the capability of CCSR, i.e., generating various SR details. Hence, we refer to the inner view consistency of NeRF and use the adjacent train views to guide the SR results. 
We introduce two very adjacent high-resolution views, which are denoted as $V_1, V_{-1}$, as the train views and optimize the interpolating view $V_0$ to see how similar the SR result of $V_0$ is to the ground truth HR view.
If the output of CCSR is the same as the HR ground truth, we believe the CCSR is capable of generating any SR solutions under guidance.
We investigate the output of CCSR at view $V_0$ with PSNR. 
The ablation study provides a new evaluation to see the capacity of generating various HR results for all the NeRF super-resolution problems.

As shown in Tab.~\ref{tab:quan_psnr}, Super-NeRF achieves the best performance of PSNR even compared to NeRF-SR with ground truth finetune, which demonstrates that Super-NeRF can indeed generate the HR ground truth result guided by the HR adjacent views.  The various SR generation capacity of the CCSR is proven.

\begin{table}
\centering
\caption{Quantitative comparison between Super-NeRF and the baselines on PSNR. }
\begin{tabular}{ccc}
\toprule 
 &LLFF&Synthetic 360\\
\midrule 
LR-NeRF&25.7&24.6\\
ESR-NeRF&26.2&23.7\\
NeRF-SR(w/o finetune)&25.6&27.1\\
NeRF-SR(with finetune)&26.3&27.9\\
Super-NeRF&\textbf{33.2}&\textbf{31.3}\\
\bottomrule
\end{tabular}
\label{tab:quan_psnr}
\end{table}

\begin{figure}[t]
\centering
  \includegraphics[width=0.45\textwidth]{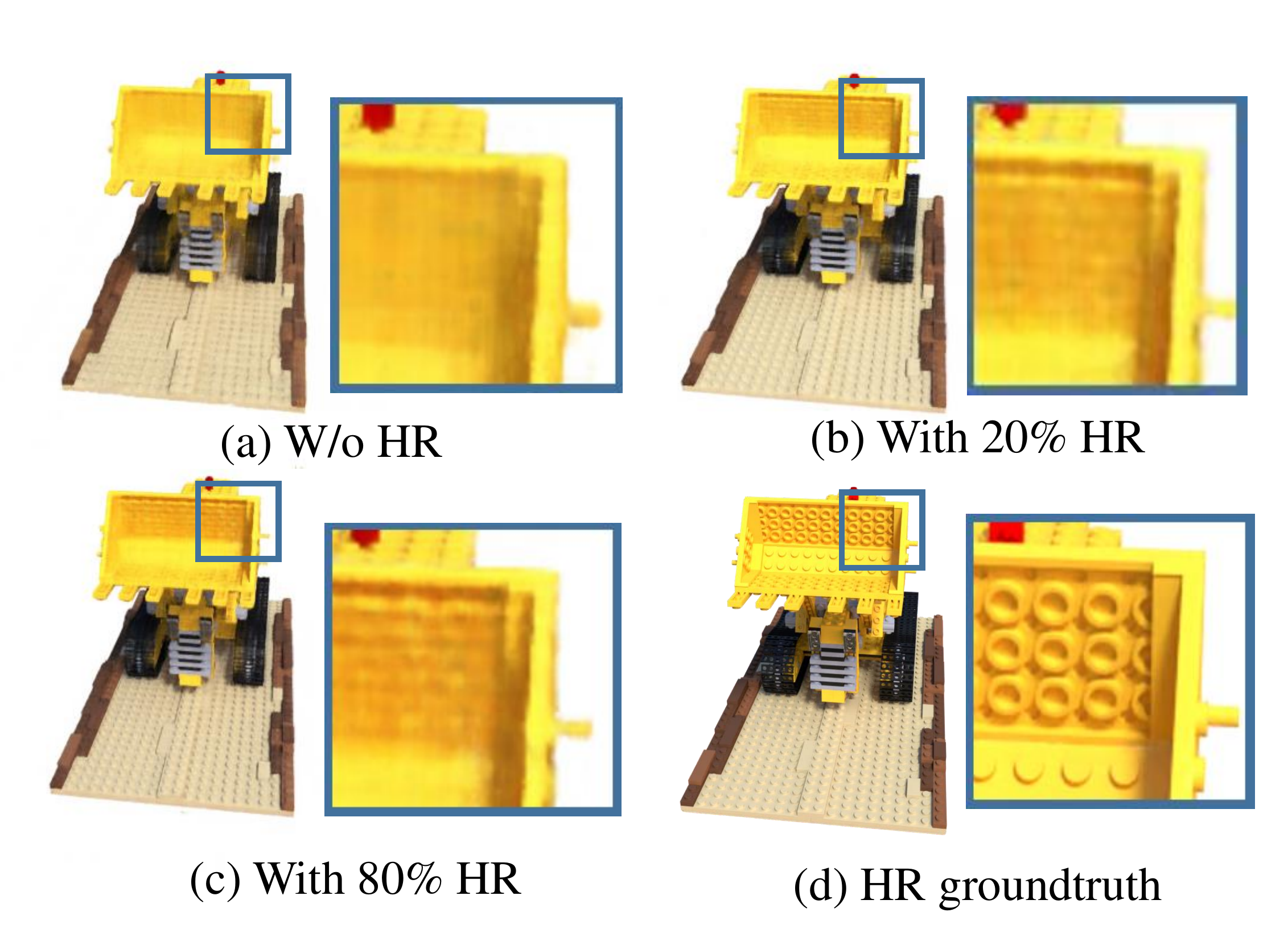}
  \caption{Performance comparison with total LR views (a), $20\%$ HR ground truth views (b), with $80\%$ ground truth views (c), and ground truth view. }
  \label{fig:LH}
\end{figure}

Further, we construct a mixed-resolution setup: given different percentages of HR ground truth viewpoints, we investigate the image quality of novel views in the Lego scene. We conduct the  comparison with total LR views in Fig.~\ref{fig:LH}(a), $20\%$ HR ground truth views in Fig.~\ref{fig:LH}(b), $80\%$ ground truth views in Fig.~\ref{fig:LH}(c), as well as the HR ground truth in Fig.~\ref{fig:LH}(d).

\begin{figure}[t]
 \centering
  \includegraphics[width=0.45\textwidth]{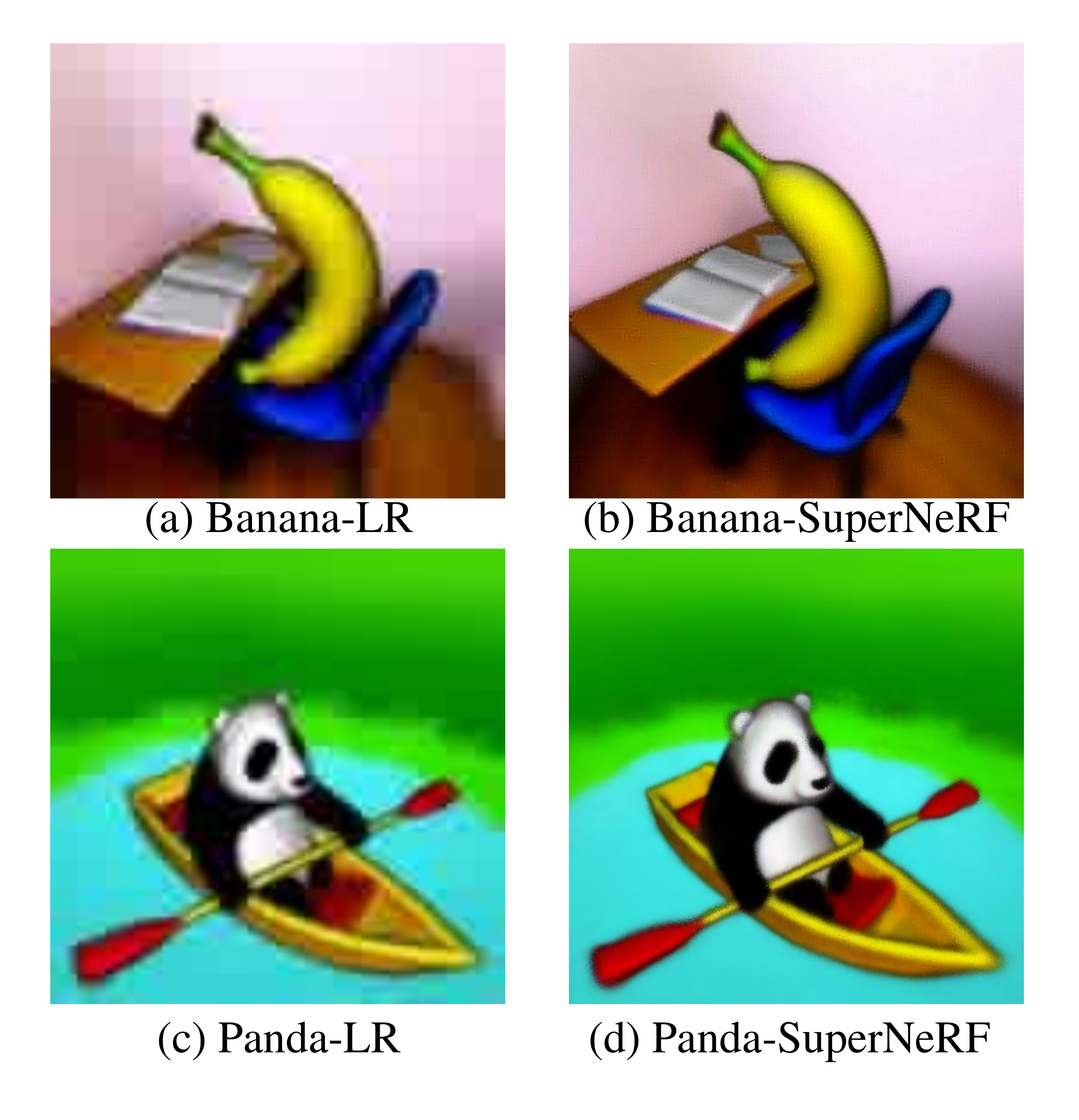}
  \caption{Super-NeRF performance in Dreamfusion\cite{dreamfusion} dataset. Our Super-NeRF successfully demonstrates good SR performance on AI-generated NeRFs. Please zoom in for clear visualization. }
  \label{fig:DF}
\end{figure}

Based on the results, we can see that the proposed Super-NeRF is robust to the hybrid resolution input.  With the increase of high-resolution viewpoints, the Super-NeRF tends to generate HR details consistent with the HR ground truth. The experiment of the hybrid resolution verifies the scalability of Super-NeRF. In the practical implementation, the Super-NeRF can be conducted in the hybrid resolution of the image capture system, which extends the practical feasibility in cross-device tasks.

\subsection{Generality in 3D AIGC}

We extend the proposed Super-NeRF on the Dreamfusion\cite{dreamfusion} data, which contains the 3D scenes generated by AI. 
Dreamfusion proposes a transformation method from 2D text to 3D fields and accomplishes the remarkable demo of a 360-degree synthesis scene. We choose several groups of Dreamfusion datasets as the input to the Super-NeRF to investigate the SR capacity in the AI-generating data. The results of the banana scene and panda scene are shown in Fig.~\ref{fig:DF}. Fig.~\ref{fig:DF}(a)(c) denotes the LR images of the different scenes in DreamFusion, Fig.~\ref{fig:DF}(b)(d) denotes the corresponding SR results of Super-NeRF. According to the result, the image quality that Super-NeRF reconstructs is improved compared to the input. Note that we first downsample the AI-generated images to a similar level to our NeRF super-resolution input. Although the scene is quite simple, Super-NeRF has the ability to improve AI-generated content. We believe that with the improvement of AI generation and also 2D SR backbones, Super-NeRF can provide more significant improvements.

\section{Conclusion and Future works}
\label{co}
\noindent{\bf Limitations:} In this paper, we only verify the Super-NeRF based on a $4\times$ SR model as the backbone. According to the extensive structure of Super-NeRF, in future work, we consider increasing the capacity of Super-NeRF to $8\times$ and $16\times$ by using more lightweight and strong 2D SR backbones. 
Also, the training is relatively slow due to the vanilla NeRF backbone we used, using the SOTA NeRF networks such as TensorRF~\cite{tensorrf} or InstantNGP~\cite{instantngp} will significantly improve the overall efficiency without changing the overall framework and the training strategy. 

{\bf Conclusion:} In this paper, we present Super-NeRF, a NeRF super-resolution method with LR input only. As far as we know, this is the first work achieving the generative NeRF super-resolution task considering the view consistency in the 3D space. We leverage the mutual learning strategy on the iterative training of both the proposed CCSR and NeRF, leading to the SR solution with view-consistency regularization.
We believe the significant SR performance and strong generalization capacity of the Super-NeRF will benefit both future research and practical applications.

{
\bibliographystyle{ieee_fullname}
\bibliography{egbib}
}

\end{document}